\documentclass[letterpaper, 10 pt, conference]{ieeeconf}  % Comment this line out if you need a4paper
\IEEEoverridecommandlockouts                              % This command is only needed if 
\usepackage{booktabs}

\usepackage{amsfonts,amssymb}  
\usepackage[cmex10]{amsmath}
\usepackage[pdftex]{graphicx}
\usepackage{array,cite,url}
\usepackage{bm}
\usepackage{float}

\newtheorem{prob}{\bf Problem}
\newtheorem{df}{\bf Definition}

\newcommand{\pder}[2]{\ensuremath{\frac{\partial #1}{\partial #2}}}

\title
{\LARGE \bf A Hybrid Compositional Approach to Optimal Mission Planning for Multi-rotor UAVs using Metric Temporal Logic
}

\author{Usman A. Fiaz and John S. Baras
\thanks{The authors are with the Department of Electrical and Computer Engineering and the Institute for Systems Research at University of Maryland, College Park, 20742 MD, USA. \tt\small $($\{fiaz,baras\}@umd.edu$)$}
}%
\begin{document}

\maketitle
\thispagestyle{empty}
\pagestyle{empty}
%\vspace{-0.5cm}
%----------------------------------------------------------------------------------------------------------------------------------------------------------
%----------------------------------------------------------------------------------------------------------------------------------------------------------

\begin{abstract}       % Abstract of not more than 250 words.
This paper investigates a hybrid compositional approach to optimal mission planning for multi-rotor Unmanned Aerial Vehicles (UAVs). We consider a time critical search and rescue scenario with two quadrotors in a constrained environment. Metric Temporal Logic (MTL) is used to formally describe the task specifications. In order to capture the various modes of UAV operation, we utilize a hybrid model for the system with linearized dynamics around different operating points. We divide the mission into several sub-tasks by exploiting the invariant nature of various task specifications i.e., the mutual independence of safety and timing constraints along the way, and the different modes (i,e., dynamics) of the robot. For each sub-task, we translate the MTL formulae into linear constraints, and solve the associated optimal control problem for desired path using a Mixed Integer Linear Program (MILP) solver. The complete path is constructed by the composition of individual optimal sub-paths. We show that the resulting trajectory satisfies the task specifications, and the proposed approach leads to significant reduction in computational complexity of the problem, making it possible to implement in real-time.

\end{abstract}
%
%%\vspace{0.1cm}
%\begin{keywords}
%Modular robots, programmable self-assembly, decentralized self-reconfiguration, autonomous systems.
%\end{keywords}

%----------------------------------------------------------------------------------------------------------------------------------------------------------
%----------------------------------------------------------------------------------------------------------------------------------------------------------

\section{Introduction}
Multi-rotor Unmanned Aerial Vehicles (UAVs) in general and quadrotors in particular, find enormous applications in several key areas of research in academia as well as industry. These include but are not limited to search and rescue, disaster relief, surveillance, autonomous aerial transport, and entertainment. The motivation behind this expanding zeal towards multi-rotor UAVs is twofold. First, these are highly inexpensive robots as compared to their fixed-wing counterparts, and are therefore leading the way as a standard testbed in much of the ongoing research in aerial robotics and motion planning. Second, these are extremely agile robots capable of much higher maneuverability in comparison with the other UAV classes, namely fixed-wing and helicopter style UAVs. This salient feature edges them as a feasible platform to operate in congested environments as well, such as crowded city skies and constrained indoor workspace.

The enormous impact of quadrotors in autonomous search and rescue missions becomes evident during natural disasters, where it is impossible for humans and terrestrial robots to access highly cluttered spaces to rescue people from life threatening situations. In such scenarios, these agile multi-rotor UAVs come to our rescue. They can either locate and grab a target themselves or can serve as a guide for other robots as well as humans to help evacuate a target. For this reason, an enormous amount of work has been done in designing new control strategies and planning algorithms for this specific class of aerial robots, with the aim to deploy them eventually in safety and time critical missions \cite{latombe,lavalle}.

Given any high level task, it is a standard practice in classical motion planning literature, to look for a trajectory or a set of trajectories, which the robot can follow (i.e., which its dynamics permit), while satisfying the desired task specifications \cite{fiaz2019usbot}. This gives rise to the notion of optimal path planning, which considers an optimal path in the sense of optimizating some suitable cost function and a control law, to go from one position to another in space while satisfying some constraints \cite{choset}. Traditionally, several interesting methods such as potential functions \cite{xi}, and optimization through generation of probabilistic maps in high dimensional state-space \cite{Sharma, fiaz_2018}, have been used for robot mission planning. However, all these methods tend to fail in situations where the task involves some finite time constraints or dynamic specifications. Aerial surveying of areas and time-critical search and rescue scenarios are two common examples of such tasks \cite{gholamidrone}. 

Temporal logic \cite{baier, Clarke, Quottrupi} seems to address this problem, since it enables us to specify complex dynamic tasks in compact mathematical form. A bulk of modern motion planning literature is based on Linear Temporal Logic (LTL) \cite{goerzen2010survey, alur2015principles}, which is useful for specifying tasks such as visiting certain objectives periodically, surveying areas, ensuring stability and safety etc. \cite{frazzoli2002, kress2009temporal, plaku2016motion}. However, from a control theory perspective, LTL only accounts for timing in the infnite horizon sense i.e., it can only guarantee something will \emph{eventually} happen and is not rich enough to describe finite time constraints. In addition, the traditional LTL formulation such as in \cite{Pappas}, assumes a static environment, which does not admit incorporating dynamic task specifications. 

On the other hand, Metric Temporal Logic (MTL) \cite{MTL,MTL1}, can express finite time requirements between various events of the mission as well as on each event duration. This allows us to specify safety critical missions with dynamic task specifications and finite time constraints. An optimization based method for LTL was proposed in \cite{KaramanCDC}, and \cite{WolffICRA}, where they translate the LTL task specifications to Mixed Integer Linear Programming (MILP) constraints, which are then used to solve an optimal control problem for a linear point-robot model. This work was extended in \cite{zhou2016timed}, where the authors used bounded time temporal constraints using extended LTL and a timed automata approach for motion planning with linear system models. However, this approach did not incorporate a well defined or rich dynamical model of the robot, and also illustrated significant computational complexity issues for the proposed method. Similarly, optimization based methods with MTL specifications were presented in \cite{nikou2016cooperative} for single and multiple robots respectively. However, in both cases, the computation of the resulting trajectory is expensive ($\sim$500 sec), and hence cannot be performed in real-time. Moreover, these works have put high constraints on the robot maneuverability, by limiting the dynamics to a simple linear (point-robot) model, which is contradictory to the primary reason for deploying quadrotors in constrained dynamic environments. 

In this work, we propose a compositional optimization based approach for motion planning with two UAVs in a time critical search and rescue mission scenario. We use MTL specifications to formally describe the task, and unlike all of the described previous works, we utilize a complete hybrid dynamical model of the robot (i.e., the quadrotor), to capture its various modes of operation. We divide the mission into several sub-tasks based on the different dynamics of the robot, and the invariant nature of safety and timing constraints along the way. For each sub-task, we translate the MTL formulae into mixed integer linear constraints, and solve the associated optimization problem for desired optimal path using a MILP solver. The final path is generated by composing the individual optimal sub-paths. We show that by breaking down the motion planning problem into several sub-problems, we can compute the optimal sub-paths in real time while satisfying the safety and timing constraints. Thus, the main contribution of this paper is the demonstration of a real-time close-to-optimal mission planner for multiple UAVs with MTL specifications, while using a complete hybrid model for the robot with minimum and realistic assumptions.

The rest of the paper is organized as following. In Section~\ref{sec:ps}, we define the problem and describe the task using a custom built workspace. Section~\ref{sec:prelim} presents the notation used and some preliminaries on system dynamics, and MTL. In Section~\ref{sec:quad}, we define the hybrid dynamical model for the quadrotor and discuss the linearization strategy for its various modes. Section~\ref{sec:formulation} describes the formulation of the optimization problem for the task, and its decomposition into several sub-tasks. Then, in Section~\ref{sec:approach}, we detail our approach to solving the problem. Section~\ref{sec:sim} covers the results drawn from simulations. In Section~\ref{sec:discussion}, we summarize and analyze the outcomes of this work briefly, before concluding with some future prospects.

%----------------------------------------------------------------------------------------------------------------------------------------------------------
%----------------------------------------------------------------------------------------------------------------------------------------------------------
\section{Problem Statement}\label{sec:ps}

\begin{figure}
	\centering
		\includegraphics[width=8.6 cm]{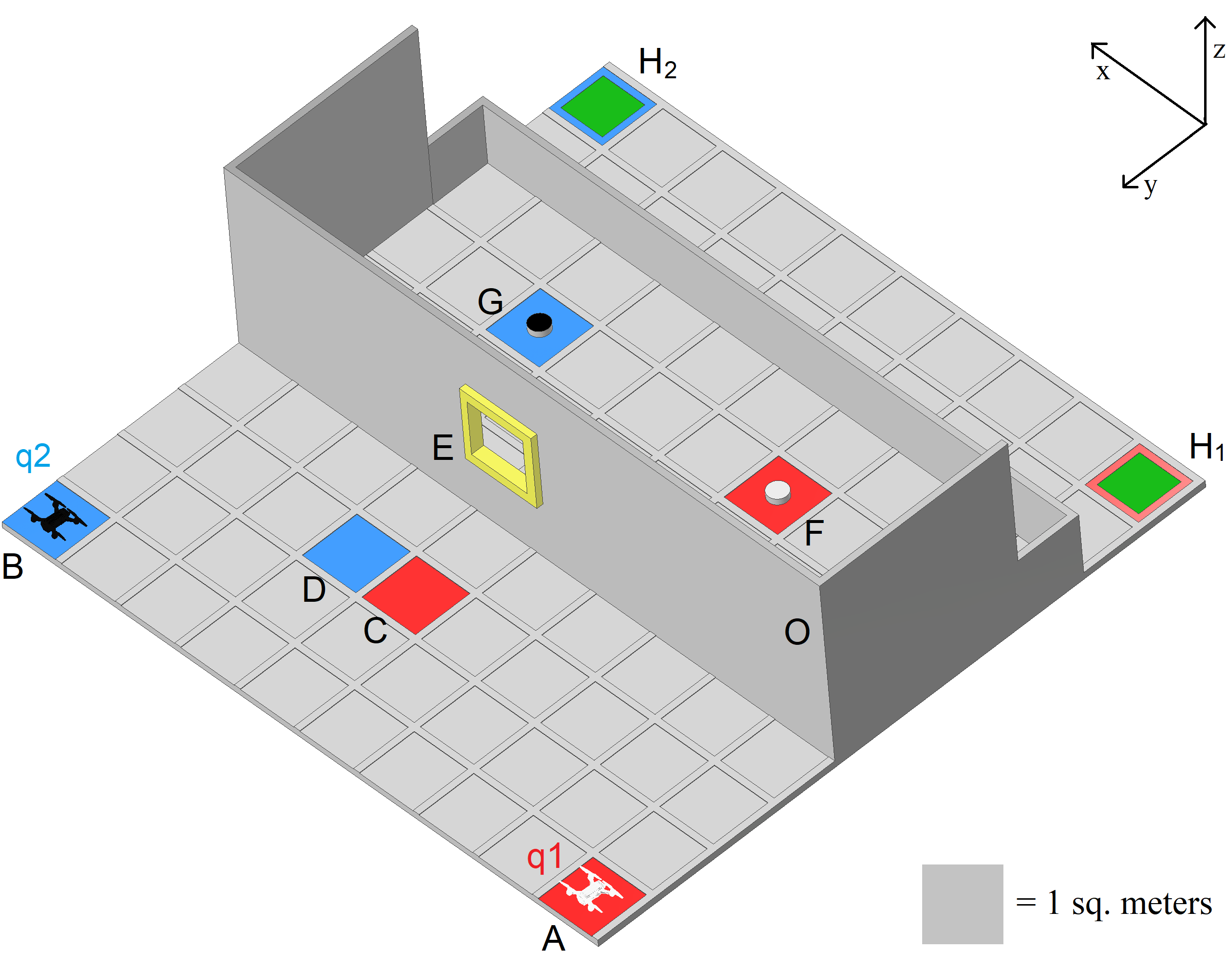}
		\caption{The computer aided design (CAD) model for the workspace used in defining the problem. The environment is a 10x10x3 cubic meter workspace which is divided into several 2D regions of interest that are labeled with alphabets, as well as marked with different colors. Red regions are reserved for quadrotor $q1$, while the blue regions are specific for quadrotor $q2$. The objects to be evacuated are assigned in the same color code as well i.e., black and white respectively for $q2$ and $q1$.}\vspace{-0.25cm}
	\label{fig:workspace}
\end{figure}

We consider a simple time critical search and rescue mission defined on a constrained workspace as shown in Fig.~\ref{fig:workspace}. The workspace is custom designed in Autodesk Inventor Professional environment, to closely suit the problem specifications. The mission is desired as follows: Starting from initial positions $A$ and $B$, two quadrotor UAVs $q1$ and $q2$ need to rescue (i.e., evacuate) two objects (shown in black and white circular disks) from a constrained environment respectively. The objects are located at $F$ and $G$ respectively, and are accessible only through a window $E$, with dimensions such that it allows only one UAV to pass at a given time. Quadrotor $q1$ has priority over quadrotor $q2$ i.e., in the case that both quadrotors arrive at the window $E$ at the same time, $q2$ has to wait for $q1$ to pass first. We also assume that there are no additional obstacles in the area other than the boundary walls $O$ and the two UAVs. The specific task for each UAV is to grasp its respective target object, and transport it to safety (marked $H_1$ and $H_2$ respectively for $q1$ and $q2$ in the workspace) in given finite time. While doing so, the UAVs need to avoid the wall $O$ as well as each other, in particular at the window $E$. In the later sections, we also use prime region notation; for example, $A'$ to represent the same 2D region $A$. Here $A'$ represents an altitude (w.r.t $z$-axis) variation of the quadrotor while it is in the same 2D region $A$.

%----------------------------------------------------------------------------------------------------------------------------------------------------------
%----------------------------------------------------------------------------------------------------------------------------------------------------------
\section{Notation and Preliminaries}\label{sec:prelim}

In this section, we describe some mathematical notation and preliminaries that are followed throughout, in rest of the paper. These are essential to set up the described problem statement as an optimal control problem with temporal constraints.

\subsection{System Dynamics}
Any general (possibly nonlinear) dynamical system can be represented in the form:
\begin{align*}
	\dot{x}(t)=f(t,x,u)
\end{align*}
where for all time $t$ continuous
\begin{itemize}
\item $x(t) \in \mathcal{X} \subseteq \mathbb{R}^n$ 
	is the state vector of the system
\item $x_0 \triangleq x(0) \in \mathcal{X}_0 \subseteq \mathcal{X}$ 
	is the initial condition of the state vector and
\item $u(t) \in \mathcal{U} \subset \mathbb{R}^m$
	is the set of control inputs which is constrained in the control set $\mathcal{U}$.
\end{itemize}
Given a nonlinear model of the system, a linearization around an operating point $(x_*(t),u_*(t))$ is expressed as:
\begin{align*}
	\dot{\hat x}(t)=A(t)\hat x(t)+B(t)\hat u(t)
\end{align*}
where for all time $t$ continuous
\begin{itemize}
\item $\hat x(t) = x(t) - x_*(t)$
\item $\hat u(t) = u(t) - u_*(t)$
\item $A(t) = \pder{f}{x}(t)\Bigr|_{x=x_*,u=u_*}$ 
\item $B(t) = \pder{f}{u}(t)\Bigr|_{x=x_*,u=u_*}$
\end{itemize}
\vspace{0.2cm}
If time $t$ is discretized, then the system dynamics take the form:
\begin{equation} \label{eqn1}
x(t+1)=f(t,x(t),u(t))
\end{equation}
where as before, $x(t) \in \mathcal{X}$, $x(0) \in \mathcal{X}_0 \subseteq \mathcal{X}$, and $u(t) \in \mathcal{U}$ for all $t=0,1,2, \cdots$. 
Let us denote the trajectory for System~(\ref{eqn1}), with initial condition $x_0$ at $t_0$, and input $u(t)$ as: $\mathbf{x}_{t_0,x_0,u(t)}=\{x(t)~|~t\geq t_0,  x(t+1)=f(t,x(t),u(t)), ~ x(t_0)=x_0\}$. However, in this paper, for the sake of convenience, we use the shorthand notation i.e., $\mathbf{x}_{t_0}$ instead of $\mathbf{x}_{t_0,x_0,u(t)}$ to represent system trajectory whenever no explicit information about $u(t)$ and $x_0$ is required.
Similar to the the continuous time case, the corresponding linearized system in discrete time looks like:
\begin{align}
\label{eq:linearDyn}
	{\hat x}(t+1)=A(t)\hat x(t)+B(t)\hat u(t)
\end{align}
 for all $t=0,1,2, \cdots$. We use System~(\ref{eq:linearDyn}) form dynamics in our final problem formulation in Section~\ref{sec:formulation}.
\subsection{Metric Temporal Logic (MTL)}
Now, we introduce some essential terminology and semantics for MTL. The convention on MTL syntax and semantics followed in this paper is the same as presented in \cite{MTL}. More details on specifying tasks as MTL formulae can also be found in \cite{MTL1}.
%\\
\begin{df}\label{defat}
\textit{An atomic proposition is a statement with regard to the state variables $x$ that is either $\mathbf{True} (\top)$ or $\mathbf{False} ( \perp)$ for some given values of $x$.} \cite{alur2015principles}
\end{df}
Let $\Pi =\{ \pi_1, \pi_2, \cdots \pi_n \}$ be 
the set of atomic propositions which labels $\mathcal{X}$ as a collection of areas of interest in the workspace, which can be possibly time varying. Then, we can define a map $L$ which labels this (possibly) time varying workspace or environment as follows:
\begin{equation} \label{fdef} \notag
L: \mathcal{X} \times \mathcal{I} \rightarrow 2^{\Pi}
 \end{equation}
where $\mathcal{I} =\{[t_1,t_2]~|~t_2  > t_1 \geq 0 \} $ and $2^{\Pi}$ denotes the power set of $\Pi$ as usual. In general, $\mathcal{I}$ represents an interval of time but it may just also represent a time instance. 
For each trajectory of System~(\ref{eqn1}) i.e., $\mathbf{x}_{t_0}$ as before, the corresponding sequence of atomic propositions, which $\mathbf{x}_{t_0}$ satisfies is given as: $\mathcal{L}(\mathbf{x}_{0})=L(x(0),0)L(x(1),1)...$.

We later specify the tasks formally using MTL formulae, which can incorporate finite timing constraints. These formulae are built on the stated atomic propositions (Definition~\ref{defat}) by following some grammar.
\begin{df} \label{def1}
 \textit{The syntax of MTL formulas are defined in accordance with the following rules of grammar:}
 \begin{center}
 $\phi ::= \top ~| ~\pi~ |~\neg \phi~ | ~\phi \vee \phi ~|~\phi \mathbf{U}_I \phi  ~ $
 \end{center} 
 \end{df}
where $I\subseteq [0, \infty]$, $\pi \in \Pi$, $\top$ and $\neg\top (=\bot)$ are the Boolean constants for $true$ and $false$ respectively. $\vee$ represents the disjunction while $\neg$ represents the negation operator. $\mathbf{U}_I$ denotes the Until operator over the time interval $I$. Similarly, other operators (both Boolean and temporal) can be expressed using the grammar imposed in Definition \ref{def1}. Some examples are conjunction ($\wedge$), always on $I$ ($\Box_I$), eventually within $I$ ($\Diamond_I$) etc. Further examples of temporal operators can be found in \cite{KaramanCDC}.

\begin{df}\label{mtlsym}
 \textit{The semantics of any MTL formula $\phi$ is recursively defined over a trajectory $x_t$ as:\\
 $x_{t} \models \pi$ iff $\pi \in L(x(t),t)$\\
 $x_{t} \models \neg \pi$ iff $\pi \notin L(x(t),t)$\\
 $x_{t} \models \phi_1\vee \phi_2$ iff $x_{t} \models \phi_1$ or $x_{t} \models \phi_2$\\
 $x_{t} \models \phi_1\wedge \phi_2$ iff $x_{t} \models \phi_1$ and $x_{t} \models \phi_2$\\
 $x_{t} \models \bigcirc \phi$ iff $x_{t+1} \models \phi$\\
 $x_{t} \models \phi_1\mathbf{U}_I \phi_2$ iff $\exists t' \in I$ s.t. $x_{t+t'} \models  \phi_2$ and $\forall$ $t'' \leq t'$,\\ $ x_{t+t''} \models \phi_1$}.
\end{df}
\vspace{0.1cm}
Thus, for instance, the expression $\phi_1 \mathbf{U}_I \phi_2$ means the following: $\phi_2$ is true within time interval $I$, and until $\phi_2$ is true, $\phi_1$ must be true. Similarly, the MTL operator $\bigcirc \phi$ means that $\phi$ is true at next time instance.  $\Box_I \phi$  means that $\phi$ is always true for the time duration or during the interval $I$, $\Diamond_I \phi$ implies that $\phi$ eventually becomes true within the interval $I$. More complicated formulas can be specified using a variety of compositions of two or more MTL operators. For example, $\Diamond_{I_1} \Box_{I_2} \phi$ suffices to the following: within time interval $I_1$, $\phi$ will be eventually true and from that time instance, it will always be true for an interval or duration of $I_2$. The remaining Boolean operators such as implication ($\Rightarrow$) and equivalence ($\Leftrightarrow$) can also be represented using the grammar rules and semantics given in Definition~\ref{def1} and Definition~\ref{mtlsym}. 
Also, similar to the convention used in Definition~\ref{mtlsym}, a system trajectory $\mathbf{x}_{t_0}$ satisfying an MTL specification $\phi$ is denoted as $\mathbf{x}_{t_0} \models \phi$. This is the general temporal constraint representation, which we use later in the optimal control problem formulation.

%-----------------------------------------------------------------------------------------------------------------------------
%----------------------------------------------------------------------------------------------------------------------------------------------------------
\section{Quadrotor Dynamics}\label{sec:quad}

We adopt the generalized nonlinear dynamics for the quadrotor model presented in \cite{Kumar}. We build a hybrid model for the system with five linear modes, namely \emph{Take off}, \emph{Land}, \emph{Hover}, \emph{Steer}, and a task specific \emph{Grasp} mode. The linearization for each mode is carried out separately about a different operating point. This enables our system to have rich dynamics with less maneuverability restrictions, while each mode still being linear. This is an important point and its significance becomes apparent once we formulate the problem and present our solution approach, since it requires all constraints in the problem to be linear. (see Section~\ref{sec:approach}). 

\subsection{General Nonlinear Model}
The dynamics of a quadrotor can be fully specified using two coordinate frames. One is a fixed earth (or world) frame, and the second is a moving body frame. Let the homogeneous transformation matrix from body frame to earth frame be $R(t)$, which is a function of time $t$. In state space representation, the quadrotor dynamics are represented as twelve states namely $[x,y,z,v_x,v_y,v_z,\phi,\theta,\psi,\omega_{\phi},\omega_{\theta},\omega_{\psi}]^T$, where $\xi=[x,y,z]^T$ and $v=[v_x,v_y,v_z]^T$ represent the position and velocity of the quadrotor respectively with respect to the body frame. $[\phi,\theta,\psi]^T$ are the angles along the three axes (i.e., roll, pitch, and yaw respectively), and $\Omega=[\omega_{\phi},\omega_{\theta},\omega_{\psi}]^T$ represents the vector containing their respective angular velocities. Under the rigid body assumptions on its airframe, the Newton-Euler formalism for quadrotor in earth frame is given by: 
\begin{align} \label{quadrotor}
 \dot{\xi} & =v  \nonumber \\
 \dot{v} & = -g\mathbf{e_3} + \frac{F}{m}R\mathbf{e_3} \\
 \dot R &=R \hat{\Omega} \nonumber \\
 \dot\Omega &= J^{-1} (-\Omega \times J\Omega + \tau)\nonumber
\end{align}
where $J$ is the moment of inertia matrix for the quadrotor, $g$ is the gravitational acceleration, $\mathbf{e_3}=[0, 0, 1]^T$, $F$ is the total thrust produced by the four rotors, and $\tau=[\tau_x,\tau_y,\tau_z]^T$ is the torque vector, whose components are the torques applied about the three axes. So, $F$, $\tau_x$, $\tau_y$, and $\tau_z$ are the four control inputs to System~\ref{quadrotor}.

\subsection{Hybrid Model with Linear Modes}

\begin{figure}
	\centering
		\includegraphics[width=8.6 cm]{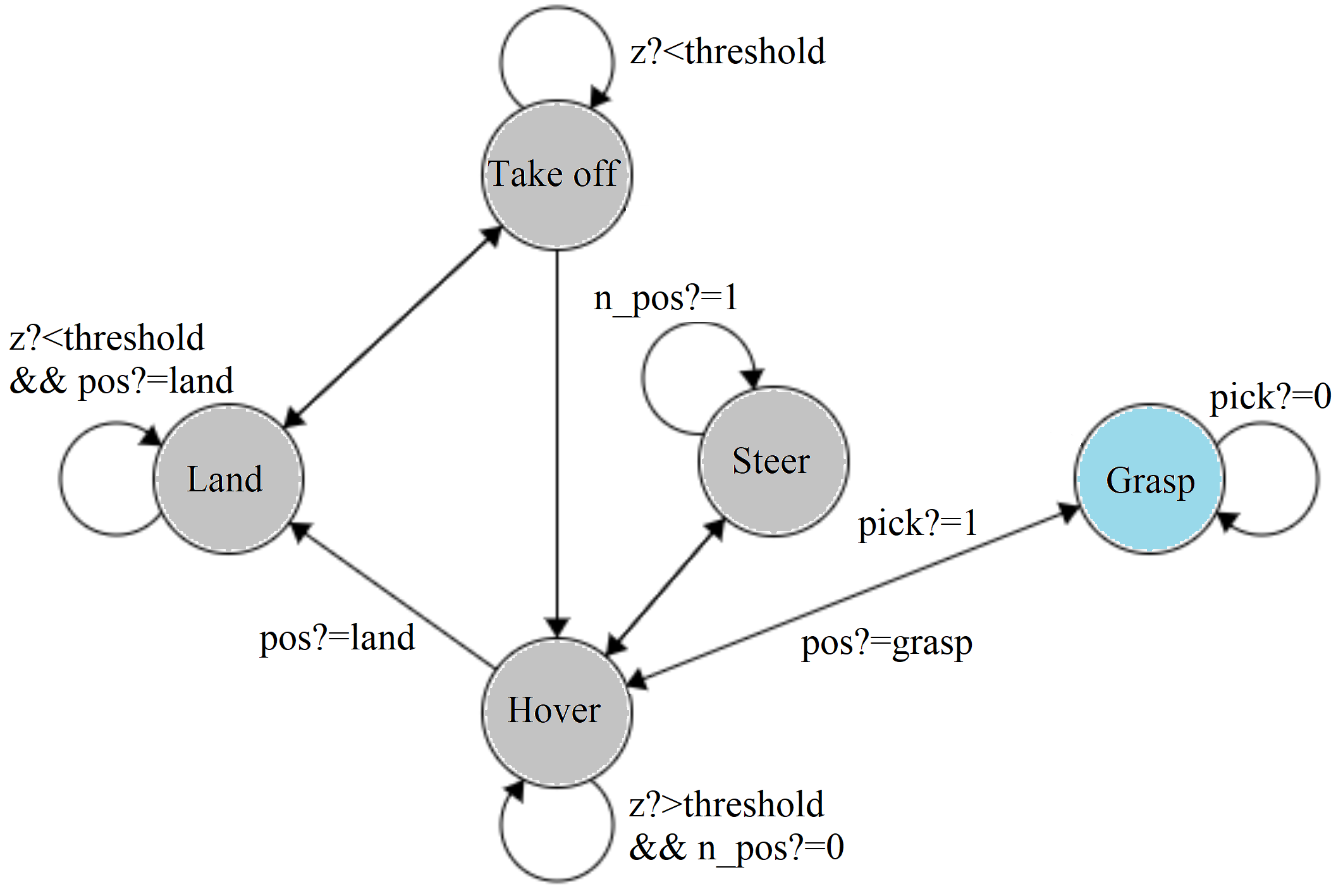}
		\caption{The simplified hybrid dynamical model for the quadrotor. Some guard conditions are hidden for readability. We use linearized dynamics around different operating points for each mode. This makes the model rich in dynamics as well as linear at the same time.}\vspace{-0.3cm}
	\label{fig:hybrid}
\end{figure}

System~(\ref{quadrotor}) serves as a starting point for for generating a hybrid model for the quadrotor with five modes, which are represented by a labeled transition system as shown in Fig.~\ref{fig:hybrid}. As usual, the states (or modes) denote the action of the robot, such as \emph{Take off} and \emph{Steer}, while the edges represent the change or switch to another action. The change is governed by some suitable guard condition. Note, that some edges donot exist; for example, the quadrotor cannot go from \emph{Land} to \emph{Hover} without taking the action \emph{Take off}. Each state or mode of the transition system follows certain dynamics, which result from a linearization of System~(\ref{quadrotor}) around a different operating point. As an example, consider the \emph{Hover} mode. One possible choice of operating points for the linearization of system dynamics in this mode is $\psi = 0$. This implies that the two states $\psi$ and $\omega_{\psi}$ i.e., the yaw angle and its respective angular velocity are identically zero, and thus can be removed from the state space representation. Consequently, the state space dimension is reduced to ten, and the control set is reduced to three inputs as well; i.e., $F, \tau_x$, and $\tau_y$. The resulting linearized model can be written in standard (discrete time) form as: ${\sigma(t+1)} = A(t)\sigma(t) + B(t)\gamma(t)$, where $\sigma(t)$ is the state, and $\gamma(t)$ is the input (in vector notation), with the two system matrices given as:
\vspace{0.25cm}\\
 $A=\begin{bmatrix} \mathbf{0} & I & \mathbf{0} & \mathbf{0} \\ 
 \mathbf{0} & \mathbf{0} & \begin{bmatrix} 0  & g \\ -g & 0\\0 & 0 \end{bmatrix} & \mathbf{0}\\ 
 \mathbf{0} &\mathbf{0} &\mathbf{0} & I \\ \mathbf{0} &\mathbf{0} &\mathbf{0} &\mathbf{0}\\
 \end{bmatrix}; B = \begin{bmatrix} \mathbf{0} & \mathbf{0}\\ \begin{bmatrix} 0\\  0 \\ 1/m \end{bmatrix} &\mathbf{0} \\ \mathbf{0} & \mathbf{0} \\\mathbf{0} & I_{2 \times 3} J^{-1}\\ \end{bmatrix}$
where $I_{2,3}= [I_{2,2}~~\mathbf{0}_{2,1}]$, and all zero and identity matrices in $A(t)$ and $B(t)$ are of proper dimensions. We adopt similar procedure to linearize System~(\ref{quadrotor}) around other operating points for different modes, and obtain linearized dynamics for the hybrid model. Here, we omit the discussion about the selection of these operating points, but it can be found in \cite{Kumar} and \cite{Garcia}.

\subsection{The Grasp Mode}
\begin{figure}[H]
	\centering
		\includegraphics[width=8.0 cm]{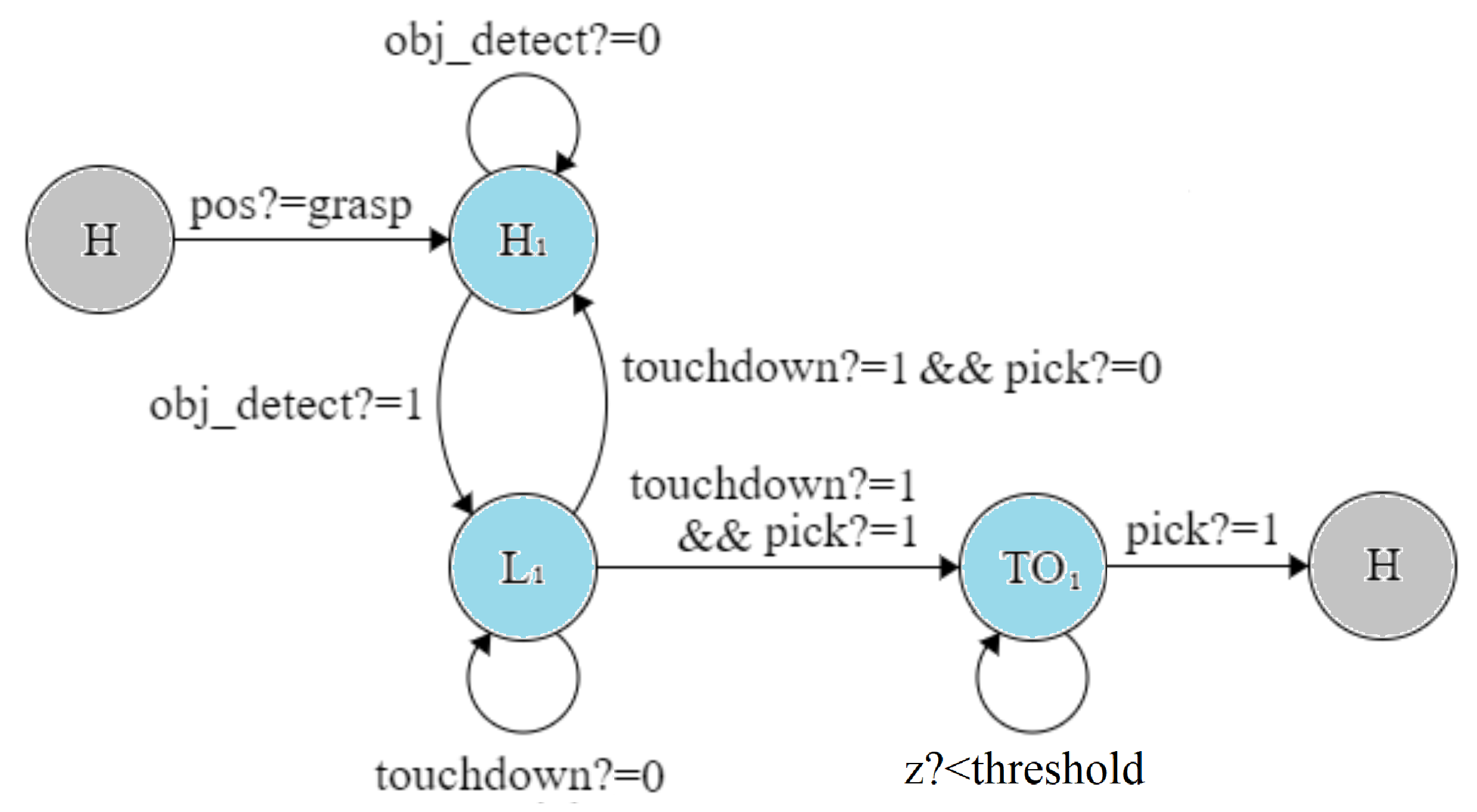}
		\caption{The \emph{Grasp} mode expressed as a combination of \emph{Hover}($H_1$), \emph{land}($L_1$), and \emph{Take off}($TO_1$) modes (colored cyan), with special guard conditions.}\vspace{-0.25cm}
	\label{fig:grasp}
\end{figure}

Grasping in general is a very challenging problem, in particular when dexterity based manipulation is involved. Since, \emph{Grasp} is the only task specific dynamical mode of our system, it was advisable to simplify the grasping routine within the high level task. Thankfully, in case of aerial grasping, some modern passive aerial grasping mechanisms \cite{fiaz2017passive} have been shown to be very reliable in grasping an object with an instantaneous touchdown onto its surface \cite{fiaz2018intelligent}. Under this instantaneous touchdown and grasp assumption \cite{fiaz2017thesis}, we can express the \emph{Grasp} mode as a switching combination of \emph{Hover}, \emph{Land}, and \emph{Take off} dynamics with special guard conditions. This clearly simplifies the problem of having the need to introduce a complex gripper and its end-effector dynamics into the \emph{Grasp} mode. Figure~\ref{fig:grasp} depicts this representation of \emph{Grasp} mode in terms of the \emph{Hover}, \emph{Land}, and \emph{Take off} dynamics.

%\begin{figure}
%	\centering
%		\includegraphics[width=8.6 cm]{PickingSM.pdf}
%		\caption{The complete \emph{grasp} mode state machine. The key idea is }
%	\label{fig:graspSM}
%\end{figure}
%---------------------------------------------------------------------------------------------------------------------------------------------------------
%----------------------------------------------------------------------------------------------------------------------------------------------------------
\section{Problem Formulation}\label{sec:formulation}

The described search and rescue problem can be set up as a standard optimal control formulation in discrete time. Given some general system dynamics (\ref{eqn1}), the objective is to find a suitable control law that steers each quadrotor through some regions of interest in the workspace within desired time bounds, so that it evacuates the target safely. This control also optimizes some cost function, while the associated task constraints are specified by an MTL expression. Let $\phi$ denote the MTL formula for a task specification, and $J(x(t,u(t)),u(t))$ be the cost function to be minimized. For the stated search and rescue scenario, a possible MTL task specification for quadrotor $q1$ can be: 
\[\phi = \Diamond_{[0,T_1]}F \wedge \Box_{[0,T_2]}F' \wedge \Diamond_{[0,T_3]}\Box H_1 \wedge \Box \neg O \wedge \Box\neg q2
\] 
where $T_3 > T_2 > T_1$ are discrete time units. A similar specification can be written for quadrotor $q2$. Then, the corresponding optimization problem is given as:
\begin{prob}\label{problem1}
\[\begin{array}{c c} \underset{x,u}{\min} \text{ }   &~J(x(t,u(t)),u(t)) \\ \text{subject to  } &x(t+1)=f(t,x(t),u(t)) \\ &\mathbf{x}_{t_0} \models \phi\\ \end{array}\]
\end{prob}\vspace{0.25cm}
Problem~\ref{problem1} is a discrete time optimal control problem with non-linear system dynamics, which is difficult to solve even in cases with simple $\phi$. In our particular case, it is only harder. Thus, a standard practice is to use linear dynamics at the expense of sacrificing maneuverability of the robot, in order to simplify the problem (as is done in \cite{zhou2015optimal} etc.).
\begin{prob}\label{problem2}
\[\begin{array}{c c} \underset{x,u}{\min} \text{ }   &~J(x(t,u(t)),u(t)) \\ \text{subject to  } &x(t+1)=At)x(t)+B(t)u(t) \\ &\mathbf{x}_{t_0} \models \phi\\ \end{array}\]
\end{prob}\vspace{0.25cm}
However, since we are using a hybrid model for the system, and not a linearization about a single operating point only, we are not sacrificing the dynamics and hence the maneuverability of the robot as much. Indeed we have five different modes or sets of linear dynamics associated with the system. Therefore, now the question arises that given an MTL task specification $\phi$, what linear dynamics or which mode is best suited to solve Problem~\ref{problem2}. The answer to this question leads us to a very interesting conclusion. In general, the task specification $\phi$ is complex and asks the robot to perform actions which require the use of multiple modes. This is true in our problem as well. Therefore, an efficient way to solve Problem~\ref{problem2} is to break down the original task into several sub-tasks by exploiting their invariant properties along the way, and based on the various modes associated with each sub-task. 

Thus, having a hybrid dynamical model with linear modes (which are linearizations of System~(\ref{quadrotor}) at different operating points) enables us to break down the task specification $\phi$ into several sub-tasks. In addition, a closer inspection of $\phi$ reveals that all safety and timing constraints specified by $\phi$ need not be satisfied by either of the robots during the whole mission. There are some critical safety constraints such as "always avoid $O$" that need to be satisfied most of the times, but majority of the constraints are purely local, based on the position of the robot in the workspace. For example, a timing constraint for quadrotor $q1$ to grasp the object at location $F$ within 10 time units is independent from (or invariant of) a timing constraint for it to reach location $C$ from $A$ within 5 time units, and so on. The mutual invariance of such local constraints or sub-tasks can also be formally verified using UPAAL \cite{behrmann2006uppaal} for instance.

Thus, Problem~\ref{problem2} can be replaced by a collection of smaller optimization problems, each with a sub-task specification represented as an MTL formula $\phi_{sub_i}$, with the associated $ith$-mode linear dynamics from the hybrid model. Further, we choose the cost function to be linear as well i.e., $\sum_{t=0}^{N}|u_i(t)|$, where $N$ is the horizon for the optimal trajectory. Thus, our final formulation of the problem is given as:

\begin{prob}\label{problem3}
\[\begin{array}{c c} \underset{x_i,u_i}{\min} \text{ }   &~\sum_{t=0}^{N}|u_i(t)| \\ \text{subject to  } &x_i(t+1)=A_i(t)x_i(t)+B_i(t)u_i(t) \\ &\mathbf{x}_{t_0^i} \models \phi_{sub_i}\\ \end{array}\]
\end{prob}\vspace{0.25cm}
where $\phi_{sub_i}$ is the MTL specification for the $ith$ sub-task, $A_i(t)$, $B_i(t)$ are the linear system matrices for the $ith$ mode, and $\mathbf{x}_{t_0^i}$ is the resulting optimal trajectory for the sub-task $\phi_{sub_i}$. For example, for quadrotor $q1$, one sub-task is to go from $A$ to $C$ in 5 time units. The MTL specification for this sub-task is given by $\phi_{q1(AC)} = \Diamond_{[0,5]} C \wedge \Box \neg O$, and the associated dynamics are selected from the \emph{Steer} mode. 
%---------------------------------------------------------------------------------------------------------------------------------------------------------
%----------------------------------------------------------------------------------------------------------------------------------------------------------
\section{Solution Approach}\label{sec:approach}

Having setup Problem~\ref{problem3}, i.e., an optimization problem with linear cost function and linear dynamics, we now describe an approach to translate the MTL specification $\phi_{sub_i}$ to linear constraints. Our method is based on the approach presented in \cite{KaramanCDC} where the authors translate LTL specifications to linear constraints. We will start with a simple temporal specification and work through the procedure to convert it to mixed integer linear constraints. We will then use this example as a foundation for translating other MTL operators to equivalent linear constraints.

Consider the constraint that a trajectory $x(t)$ lies within a convex polytope $\mathcal{K}$ at time $t$. Since $\mathcal{K}$ is convex, it can be represented as an intersection of a finite number of halfspaces. A halfspace can be represented as set of points, $\mathcal{H}_i=\{ x: ~ h_i^Tx\leq a_i\}$. Thus, $x(t) \in \mathcal{K}$ is equivalent to $x(t) \in \cap_{i=1}^n \mathcal{H}_i =\cap_{i=1}^n \{ x: ~ h_i^Tx \leq a_i \}$. So, the constraint  $x(t) \in \mathcal{K}$ $\forall$ $t \in \{t_1, t_1+1, \cdots t_1+n \}$ can be represented by the set of linear constraints $\{ h_i^Tx(t)\leq a_i \}$, $\forall$ $ i =\{1,2,\cdots, n\}$ and $\forall t\in \{t_1, t_1+1, \cdots t_1+n \}$.
 
In a polytopic environment, atomic propositions (see Definition~\ref{def1}), $p,q \in \Pi$, are related to system state via conjunction and disjunction of linear halfspaces \cite{KaramanCDC}. Let us consider the case of a convex polytope and let $b_i^t \in \{0,1\}$ be some binary variables associated with the corresponding halfspaces $\{x(t) : h_i^T x(t)\leq a_i\}$ at time $t=0,...,N$. We can then force the constraint: $b_i^t=1$ $\iff$ $h_i^T x(t) \leq a_i$, by introducing the following linear constraints:
\begin{equation} 
 h_i^T x(t) \leq a_i + M(1-b_i^t) 
\end{equation} 
\[ h_i^T x(t) \geq a_i -M b_i^t +\epsilon\]
where $M$ and $\epsilon$ are some large and small positive numbers respectively. If we denote $K_t^\mathcal{K}=\wedge_{i=1}^n b_i^t$, then $K_t^{\mathcal{K}}=1$ $\iff$ $x(t) \in \mathcal{K}$. This approach is extended to the general nonconvex case by convex decomposition of the polytope. Then, the decomposed convex polytopes are related using disjunction operators. Similar to conjunction, as is described later in this section, the disjunction operator can also be translated to mixed integer linear constraints.

Let $S_\phi(x, b, u, t)$ denote the set of all mixed integer linear constraints corresponding to a temporal expression $\phi$. Using the described procedure, once we have obtained $S_p(x,b,u,t)$ for atomic propositions $p \in \Pi$, we can formulate $S_\phi (x,b,u,t)$ for any MTL formula $\phi$. 
Now, for the Boolean MTL operators, such as $\neg$, $\wedge$, $\vee$, let $t \in \{0,1,...,N\}$, and as before $K_t^\phi \in [0,1]$ be the continuous variables associated with the formula $\phi$ generated at time $t$ with atomic propositions $p \in \Pi$. Then 
$\phi=\neg p$ is the negation of an atomic proposition, and it can be modeled as:
	\begin{align} \label{nega} 
	K_t^{\phi} = 1- K_t^p
	\end{align}
 the conjunction operator, $\phi = \wedge^m_{i=1}p_i$, is modeled as:
	\begin{align}
	K_t^{\phi} & \leq K_t^{p_i}, \quad i=1,...,m \\ \nonumber 
	K_t^{\phi} & \geq 1-m+ \sum_{i=1}^m {K_t^{p_i}} 
	\end{align}
and the disjunction operator, $\phi = \vee^m_{i=1}p_i$, is modeled as:
	\begin{align}
	K_t^{\phi} & \geq K_t^{p_i}, \quad i=1,...,m \\ \nonumber 
	K_t^{\phi} & \leq \sum_{i=1}^m {K_t^{p_i}}
	\end{align}
Similar to binary operators, temporal operators such as $\Diamond, \Box$, and $\mathcal{U}$ can be modeled using linear constraints as well. Let $t \in \{0,1,...,N-t_2\}$, where $[t_1,t_2]$ is the time interval used in the MTL specification $\phi$. Then, 
eventually operator: $\phi=\Diamond_{[t_1,t_2]} p$ is modeled as:
	\begin{align}
	K_t^{\phi} & \geq K_\tau^{p}, \quad \tau \in \{t+t_1,...,t+t_2 \} \\ \nonumber
	K_t^{\phi} & \leq \sum_{\tau=t+t_1}^{t+t_2} {K_\tau^{p}} 
	\end{align}
and always operator: $\phi=\Box_{[t_1,t_2]} p$ is represented as:
	\begin{align}
	K_t^{\phi} & \leq K_\tau^{p}, \quad \tau \in\{t+t_1,...,t+t_2\}\\ \nonumber
	K_t^{\phi} & \geq \sum_{\tau=t+t_1}^{t+t_2} {K_\tau^{p}} -(t_2-t_1)
	\end{align}
and until operator: $\phi=p~\mathcal{U}_{[t_1,t_2]}~q$ is equivalent to:
	\begin{align}\label{eq:until}
	c_{tj} & \leq K_q^{j} \quad j \in \{t+t_1,\cdots,t+ t_2\}\nonumber \\ \nonumber
	c_{tj} & \leq K_p^{l} \quad l \in \{t, \cdots, j-1\}, j \in \{t+t_1,\cdots,t+ t_2\}\\ \nonumber
	c_{tj} & \geq K_q^{j} + \sum_{l=t}^{j-1} K_p^l -(j-t) \quad j \in \{t+t_1,\cdots,t+ t_2\}\\
	c_{tt} & = K_q^t  \\ \nonumber
	K_t^{\phi} & \leq \sum_{j=t+t_1}^{t+t_2}c_{tj} \\ \nonumber
	K_t^{\phi} & \geq c_{tj} \quad j \in \{t+t_1,\cdots, t+t_2\} \nonumber
	\end{align}
The equivalent linear constraints for until operator (\ref{eq:until}) are constructed using a procedure similar to \cite{KaramanCDC}. The modification for MTL comes from the following result in \cite{MTL}.
\[K_t^{\phi}= \bigvee_{j=t+t_1}^{t+t_2}\left((\wedge_{l=t}^{l=j-1}K_p^l)  \wedge K_q^j \right). \]  

All other combinations of MTL operators for example, eventually-always operator: $\phi=\Diamond_{[t_1,t_2]} \Box_{[t_3,t_4]}p$ and always-eventually operator: $\phi=\Box_{[t_1,t_2]}\Diamond_{[t_3,t_4]} p$ etc., can be translated to similar linear constraints using (\ref{nega})-(\ref{eq:until}). In addition to the collective operator constraints, at the end, we need the constraint $K_0^\phi=1$, which suffices to the overall satisfaction of a task specification $\phi$. Further details on this method can be found in \cite{KaramanCDC}. 

Using this approach, we translate the sub-task MTL constraints i.e., $\phi_{sub_i}$ to a set of mixed integer linear constraints $S_\phi (x,b,u,t)$. This way Problem~\ref{problem3} converts into a Mixed Integer Linear Program (MILP), which can be solved using a solver, quite efficiently. 

Notice, that the worst case complexity of this MILP is exponential i.e., $O(2^{mN})$, where $m$ is the number of boolean variables or equivalently the number of halfspaces required to express the MTL formula, and $N$ is the discrete time horizon. This also backs our concept of decomposing the task specification $\phi$ into several simpler sub-tasks $\phi_{sub_i}$.
%---------------------------------------------------------------------------------------------------------------------------------------------------------
%----------------------------------------------------------------------------------------------------------------------------------------------------------
\section{Simulation and Results}\label{sec:sim}

We apply the described method for solving Problem~\ref{problem3} collection, in the same workspace as shown in Fig.~\ref{fig:workspace}. It is a custom built 10x10x3 cubic meter CAD environment which is readily importable across many simulation and computation software tools. In our case study, the simulation experiments are run through YALMIP-CPLEX using MATLAB interface on an Intel NuC. It is portable computer with an Intel core i7 @ 3.7 GHz CPU, an integrated Intel Iris GPU, and 16 GBs of memory. 

\begin{figure}
	\centering
		\includegraphics[width=8.6 cm]{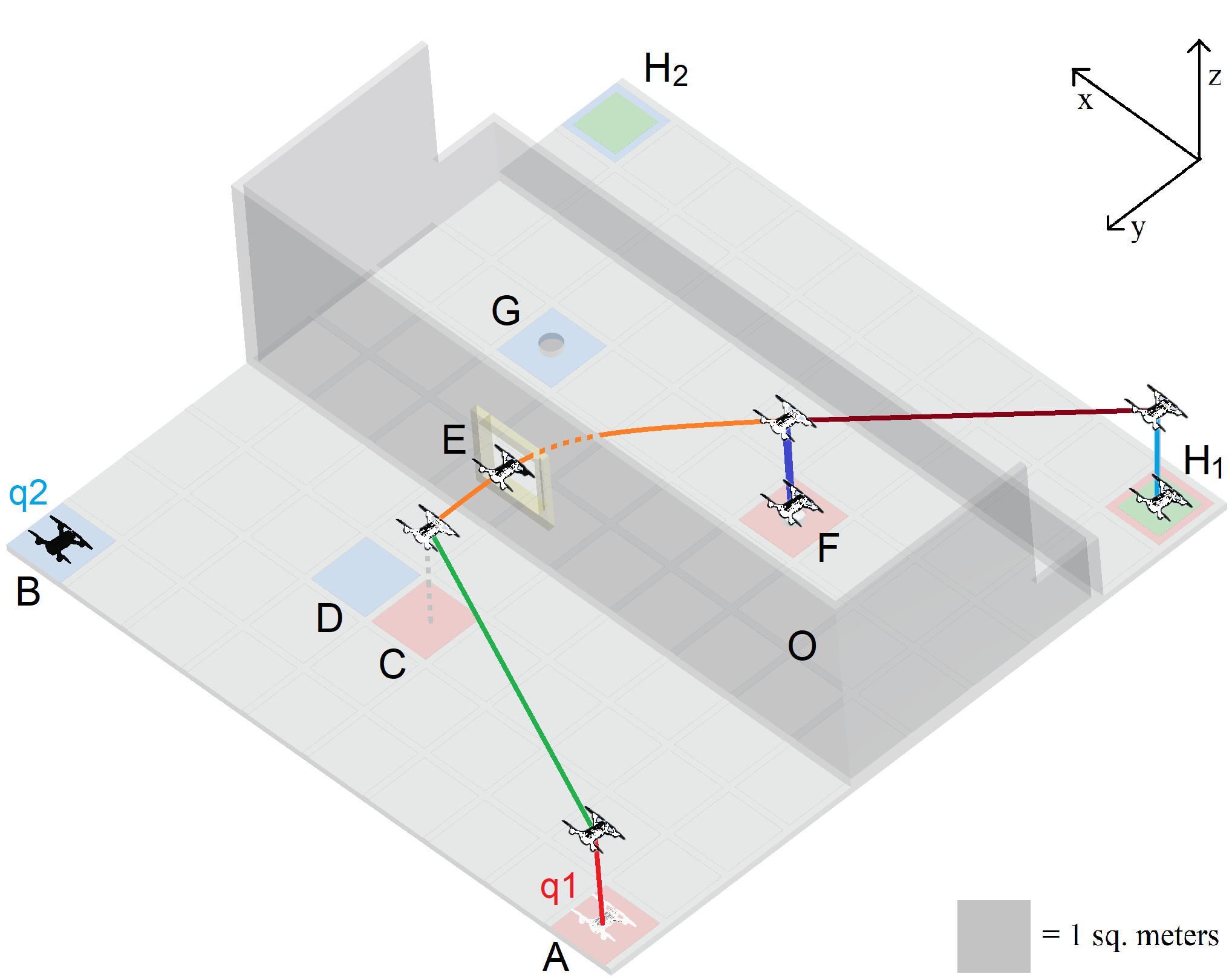}
		\caption{The resulting composed trajectories for the sub-tasks for $q1$ alone in the workspace. Optimal trajectories for sub-tasks are marked with different colors. The \emph{Hover} and \emph{Steer} altitude is set to $1.5 m$ which is the same as the $xz$-center of the window $E$. A similar plot is obtained for $q2$ only with its respective regions of interest.}\vspace{-0.25cm}
	\label{fig:q1}
\end{figure}

For the given problem statement (i.e., the task $\phi$), the sub-task specifications for the quadrotor $q1$ are specified as following along with their respective dynamics.
\[\phi_{q1(AA')} = \Box A \wedge \Diamond_{[0,5]}A' \hspace{0.5cm}  [mode: \emph{Take off}]
\] \vspace{-0.5cm}
\[\phi_{q1(AC)} = \Diamond_{[0,5]} C \wedge \Box \neg O \hspace{0.5cm} [mode: \emph{Steer}]
\]\vspace{-0.5cm}
\[\phi_{q1(CF)} = \Diamond_{[0,10]} F \wedge \Box \neg O \hspace{0.5cm} [mode: \emph{Steer}]
\]\vspace{-0.5cm}
 \[\phi_{q1(FF')} = \Box F \wedge \Diamond_{[0,10]} F' \hspace{0.5cm} [mode: \emph{Grasp}]
\]\vspace{-0.5cm}
\[\phi_{q1(FH_1)} = \Diamond_{[0,10]}H_1 \wedge \Box \neg O \hspace{0.5cm} [mode: \emph{Steer}]
\]\vspace{-0.5cm}
\[\phi_{q1(H_1H_1')} = \Box H_1 \hspace{0.5cm} [mode: \emph{Land}]
\]\\
where $O$ represents the obstacle. Using the convention defined earlier, the specification $\phi_{q1(AA')}$ requires the quadrotor $q1$ to attain desired threshold altitude (represented as $A'$) while staying inside the 2D region marked $A$. $\phi_{q1(AC)}$ requires the quadrotor $q1$ to reach $C$ within 5 time units, and $\phi_{q1(CF)}$ requires it to reach $F$ within 10 time units, while avoiding the obstacle $O$. $\phi_{q1(FF')}$ requires the robot to grasp the object at $F$ within 10 time units while staying in $F$, whereas $\phi_{q1(FH_1)}$ asks the robot to reach $H_1$ within 10 time units. Finally $\phi_{q1(H_1H_1')}$ forces $q1$ to stay at $H_1$ indefinitely.
Similarly, the sub-task specifications for quadrotor $q2$ are given as following.
\[\phi_{q2(BB')} = \Box B \wedge \Diamond_{[0,5]}B' \hspace{0.5cm}  [mode: \emph{Take off}]
\]\vspace{-0.5cm}
\[\phi_{q2(BD)} = \Diamond_{[0,5]} D \wedge \Box \neg O \hspace{0.5cm} [mode: \emph{Steer}]
\]\vspace{-0.5cm}
\[\phi_{q2(D)} = \Box D~\mathcal{U} (\neg({pos(q1)?=C}) \hspace{0.5cm}  [mode: \emph{Hover}]
\]\vspace{-0.5cm}
\[\phi_{q2(DG)} = \Diamond_{[0,10]} G \wedge \Box \neg O \hspace{0.5cm} [mode: \emph{Steer}]
\]\vspace{-0.5cm}
 \[\phi_{q2(GG')} = \Box G \wedge \Diamond_{[0,10]} G' \hspace{0.5cm} [mode: \emph{Grasp}]
\]\vspace{-0.5cm}
\[\phi_{q2(GH_2)} = \Diamond_{[0,10]}H_2 \wedge \Box \neg O \hspace{0.5cm} [mode: \emph{Steer}]
\]\vspace{-0.5cm}
\[\phi_{q2(H_2H_2')} = \Box H_2 \hspace{0.5cm} [mode: \emph{Land}]
\]\\
Notice, that quadrotor $q2$ has an additional constraint namely $\phi_{q2(D)}$, which assigns priority to $q1$ by keeping $q2$ waiting at $D$, thus ensuring that both robots do not end up at $E$ at the same time. Fairness of this specification is easy to verify, since it is impossible for $q1$ to be stuck at $C$ indefinitely. Furthermore, the \emph{Hover} mode is implicitly present in all sub-tasks and is involved in switching between different modes. However, there is no corresponding optimization required once the UAV is within the desired region. For the simulation, we use the cost function $J$ as in Problem~\ref{problem3}, with $N = 30$ i.e., the number of discrete times steps. The quadrotor dynamics for all modes are uniformally discretized at a rate of 5Hz.

\begin{figure}
	\centering
		\includegraphics[width=8.6 cm]{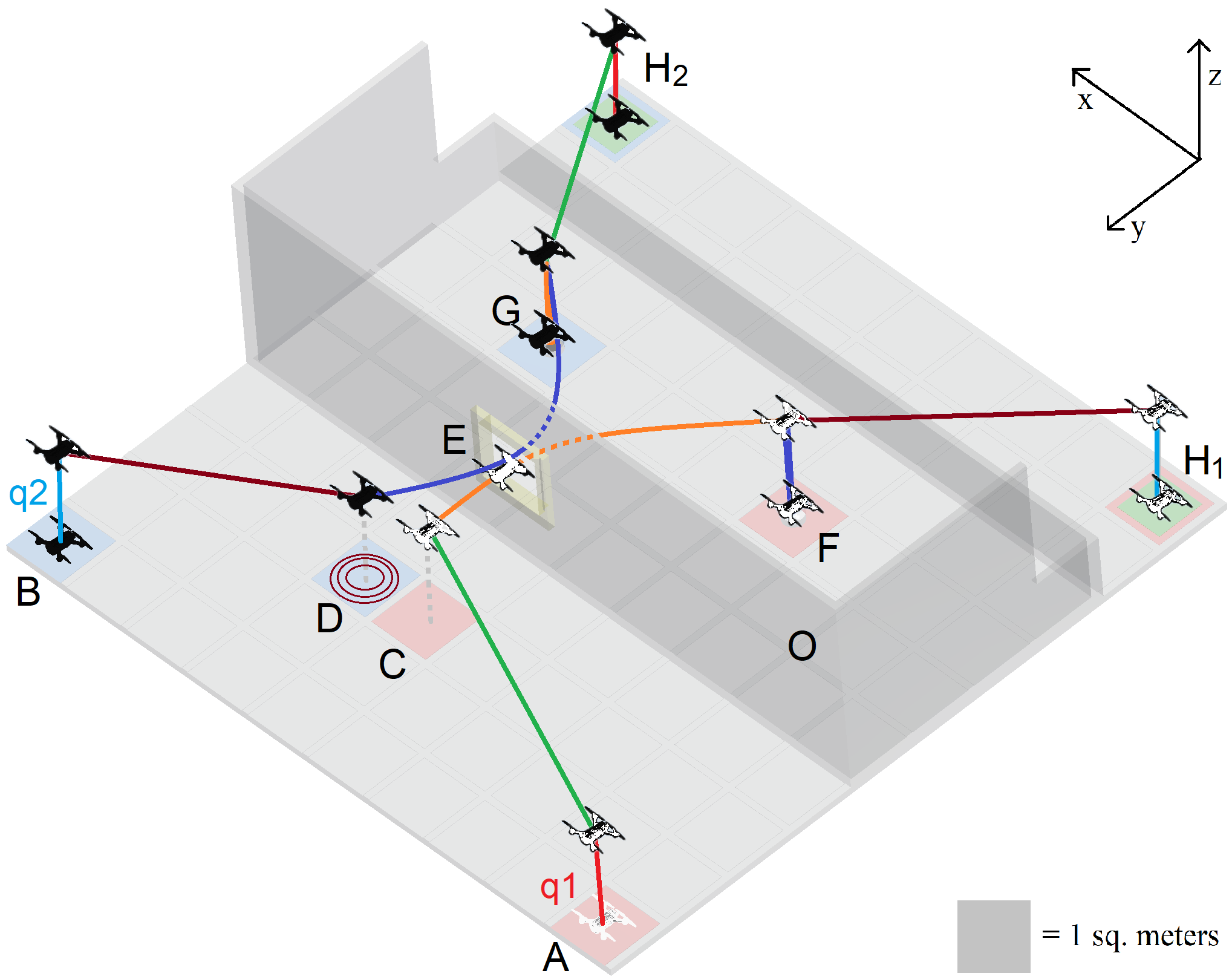}
		\caption{The resulting composed trajectories for the sub-tasks for $q1$ and $q2$ in the workspace simultaneously. Optimal trajectories for sub-tasks are marked with different colors in reverse orders for $q1$ and $q2$. The \emph{Hover} and \emph{Steer} altitude is set to $1.5 m$ as before. The number of circular rings at $D$ represent the number of time units spent by $q2$ while waiting in \emph{Hover} mode.}\vspace{-0.25cm}
	\label{fig:q1q2}
\end{figure}

The resulting trajectory for the complete mission is generated by concatenation of optimal sub-task trajectories. For quadrotor $q1$ alone, the final composed and continuous trajectory (interpolated between discrete points) is shown in Fig.~\ref{fig:q1}. The quadrotor successfully avoids the obstacle and evacuates the object to safety within specified time. In absence of $q1$, a similar trajectory for quadrotor $q2$ is obtained with its respective areas of interest. 

Figure~\ref{fig:q1q2} shows the resulting composed trajectories for both the quadrotors operating simultaneously. Again, both safely avoid the obstacle and evacuate their respective objects within given times. As expected, quadrotor $q2$ waits at $D$, for quadrotor $q1$, to avoid collision. The number of circular rings at $D$ (see Fig.~\ref{fig:q1q2}) correspond to the waiting time for $q2$, while it stays at $D$ in \emph{Hover} mode. Since, both these plots are displaying motion in 3D, it is not possible to visualize the satisfaction of timing constraints per each sub-task along a fourth axis. However, we provide some insight to the timing analysis of each sub-task for both quadrotors in terms of computation time. In addition, a closer look at the simulation data indicates that all timing constraints are indeed satisfied.

\begin{table}[t]
\caption{Computation times for sub-tasks ($\phi_{sub_i}$)}
\centering\small
\begin{tabular*}{\linewidth}{@{\extracolsep{\fill}}p{0.22\linewidth}p{0.22\linewidth}p{0.22\linewidth}p{0.22\linewidth}@{}}
\toprule
$\mathbf{Task (q1)}$ & \textbf{Time (sec)} & $\mathbf{Task (q2)}$ & \textbf{Time (sec)}\\
\midrule
$\phi_{q1(AA')}$ & 2.7 & $\phi_{q2(BB')}$ & 2.7\\
$\phi_{q1(AC)}$ & 6.3 & $\phi_{q2(BD)}$ & 5.8\\
$\phi_{q1(CF)}$ & 10.3 & $\phi_{q2(D)}$ & 2.0\\
$\phi_{q1(FF')}$ & 3.0 & $\phi_{q2(DG)}$ & 11.1\\
$\phi_{q1(FH_1)}$ & 5.7 & $\phi_{q2(GG')}$ & 3.0\\
$\phi_{q1(H_1H_1')}$ & 2.5 & $\phi_{q2(GH_2)}$ & 5.7\\
 ----------- & ----- & $\phi_{q2(H_2H_2')}$ & 2.5\\
\bottomrule
\end{tabular*}\vspace{-0.5cm}
\label{tab:computation}
\end{table}

Table~\ref{tab:computation} shows the computation times for each individual sub-task ($\phi_{sub_i}$) for both quadrotors. These numbers indicate that the proposed approach can be implemented in real-time. From an implementation point of view, the performance can be further improved by using hardware which is optimized for computation (such as Nvidia Jetson TX2 etc.).
%---------------------------------------------------------------------------------------------------------------------------------------------------------
%----------------------------------------------------------------------------------------------------------------------------------------------------------
\section{Discussion}\label{sec:discussion}

We have proposed a hybrid compositional approach to mission planning for quadrotors with MTL task specifications, and have presented an optimization based method which can be implemented in real time. Using a simple yet realistic search and rescue test case, we have demonstrated the computational efficiency of our approach, and have shown that by breaking down the task into several sub-tasks, and using a hybrid model for system dynamics, it is possible to solve the challenging problem of motion planning with rich system dynamics and finite time constraints in real time.

In addition to some promising results, this work also poses many new and interesting questions as well. For example, given a finite time constraint for the whole task, what is the best or optimal way to divide the timing constraints among various sub-tasks. Of course it is a scheduling problem, and is dependent on many factors such as robot dynamics, its maximum attainable speed, and nature of the sub-tasks as well. In our study, the individual sub-task timing constraint assignments were relaxed and uniform. However, it is worth noticing that using a complete dynamical model for the UAV puts less constraints on its maneuverability, and hence can allow it to tackle more conservative finite time constraints as well. For example, the \emph{Steer} mode in our model allows the quadrotor to achieve speeds as high as 0.5 m/s, which is not possible with the usual single mode \emph{Hover} linearization only, which involves small angle (pitch, roll) assumptions. 

The scalability features of this approach with regard to a large number of UAVs within a mission, need to be demonstrated. The invariant property of sub-tasks can also be proved rigorously using tools from formal methods. Extension of this work with more complicated tasks, dynamic obstacles, conservative constraints, and tolerances in both time and space can also yield interesting results, and are all great directions for future work.

%\section{Acknowledgment}\label{sec:ack}
%This work was partially supported by the ONR grant N00014-17-1-2622.

%\section{Supplementary Material}\label{sec:supp}
%Link to simulation code and data: https://bit.ly/2UYSOWE

\bibliography{IEEEabrv,ieeeconf}

%----------------------------------------------------------------------------------------------------------------------------------------------------------
%----------------------------------------------------------------------------------------------------------------------------------------------------------

\end{document}